\RequirePackage{fix-cm}
\documentclass[twocolumn]{svjour3}          % twocolumn
\smartqed  % flush right qed marks, e.g. at end of proof
\PassOptionsToPackage{hyphens}{url}
\usepackage[draft]{hyperref}
\usepackage{graphicx}
\usepackage{color}
\usepackage{natbib}

\newcommand{\eat}[1]{}
\begin{document}

\title{Deep Nets: What have they ever done for Vision?\thanks{For those readers unfamiliar with Monty Python see: \url{https://youtu.be/Qc7HmhrgTuQ}}}
\subtitle{}

%\titlerunning{Short form of title}        % if too long for running head

\author{Alan L. Yuille         \and
        Chenxi Liu %etc.
}

%\authorrunning{Short form of author list} % if too long for running head

\institute{Alan L. Yuille \at
              Johns Hopkins University, Baltimore, MD, USA \\
              \email{alan.l.yuille@gmail.com}
           \and
           Chenxi Liu \at
              Johns Hopkins University, Baltimore, MD, USA \\
              \email{cxliu@jhu.edu}
}

\date{Received: date / Accepted: date}
% The correct dates will be entered by the editor

\maketitle

\begin{abstract}
This is an opinion paper about the strengths and weaknesses of Deep Nets for vision. They are at the heart of the enormous recent progress in artificial intelligence and are of growing importance in cognitive science and neuroscience. They have had many successes but also have several limitations and there is limited understanding of their inner workings.  At present Deep Nets perform very well on specific visual tasks with benchmark datasets but they are much less general purpose, flexible, and adaptive than the human visual system. We  argue that Deep Nets in their current form are unlikely to be able to overcome the fundamental problem of computer vision, namely how to deal with the combinatorial explosion, caused by the enormous complexity of natural images, and obtain the rich understanding of visual scenes that the human visual achieves. We argue that this combinatorial explosion takes us into a regime where ``big data is not enough'' and where we need to rethink our methods for benchmarking performance and evaluating vision algorithms. We stress that, as vision algorithms are increasingly used in real world applications, that performance evaluation is not merely an academic exercise but has important consequences in the real world. It is impractical to review the entire Deep Net literature so we restrict ourselves to a limited range of topics and references which are intended as entry points into the literature. The views expressed in this paper are our own and do not necessarily represent those of anybody else in the computer vision community.
% \PACS{PACS code1 \and PACS code2 \and more}
% \subclass{MSC code1 \and MSC code2 \and more}
\end{abstract}

%We think it unlikely that Deep Nets in their current form will be the best long-term solution either for building general purpose intelligent machines or for understanding the mind/brain, but it is likely that many aspects of them will remain.

\section{Introduction}

In the last few years Deep Nets have enabled enormous advances in computer vision and the study of biological visual systems. But as researchers in these  areas, we have mixed feelings about them. On the one hand, we marvel at their successes and how they have led to amazing results on some real world tasks and, in academic settings, they almost always outperform alternative approaches on benchmark datasets. But, on the other hand, we are conscious of their current limitations, aware of papers \citep{DBLP:journals/cacm/Darwiche18, DBLP:journals/corr/abs-1801-00631} which criticize them from the perspectives of machine reasoning and cognitive science respectively, and are concerned about the hype that sometimes surrounds them. The nature of our research  means that we interact with research faculty in many disciplines (cognitive science, computer science, applied mathematics, neuroscience, engineering, physics, and radiology) and Deep Nets are a frequent topic of conversation. We find ourselves spending half the time criticizing Deep Nets for their limitations and the other half praising them and defending them against their critics. Not infrequently we are confidently told that ``Deep Nets can never do such and such'' (e.g., estimate 3D depth, classify objects in PASCAL without pre-training) when we already know that they have been shown to do so on benchmark datasets. This opinion paper attempts to provide a balanced viewpoint on the strengths and weaknesses of Deep Nets for studying vision, but the views expressed are our own and may not be representative of the computer vision community. Moreover, given the vast literature, the references are intended only as entry points into the literature and are far from being exhaustive. Our fundamental concern, as vision scientists and not as neural network researchers, is whether Deep Nets in their current form are sufficient to perform the vast set of visual tasks faced with the combinatorial complexity of natural images and, as we will argue, in light of the current limitations of how the community benchmarks performance.

First recall, that the purpose of vision is to know what is where by looking and can be formulated as an inverse inference problem. From the patterns of light rays that enter our eyes we can estimate many properties of the three-dimensional world.  As the authors  write these words, we can look round the room and see multiple objects (chairs, pictures, books, a robot dog, carpets, clothes, a sleeping cat, an empty wine glass, a projector, an exercise  machine, and many others). For all of them, we can estimate their shape, their position in the world, and the positions and shapes of their parts. We can also describe their properties and attributes, like the age of the carpet and how recently it has been cleaned. Humans can solve this inverse inference problem apparently effortlessly (we just open our eyes and see) but this disguises the incredible difficulty of the problem and the amount of neural resources which our brains dedicate to it. Roughly half the neurons in the human cortex are involved in visual perception, which has been called human's superpower \citep{changizi2010vision}  because of its ability to extract information about the world at a large range of distances varying from a fraction of an inch to millions of light years. The difficulty of vision became appreciated as AI researchers tried to design algorithms to mimic human visual abilities.  A few vision researchers proposed bold speculative conceptual theories of vision \citep{gibson1986ecological, gregory1973eye, marr1982vision} but most researchers in computer vision followed a more pragmatic engineering strategy, where individual visual tasks (e.g., edge detection, binocular stereo, object recognition) were studied separately. This strategy enabled computer vision researchers to make gradual progress by becoming familiar with the challenges of the specific tasks and by borrowing and adapting mathematical and computational techniques from other disciplines (e.g., mathematics, computer science, engineering, etc). A major innovation (which started slowly in the late 1980’s but accelerated rapidly after 2000) was the development of benchmark annotated datasets. These served two purposes. Firstly, they allowed researchers to quantify their algorithms and compare performance. Secondly they enabled the use of learning algorithms and led to a symbiotic relationship with the rapidly developing field of machine learning. Computer vision mostly converged to a standard paradigm where learning-based algorithms were compared on annotated benchmark datasets.  But we will argue that, in light of the combinatorial  complexity of natural images, that the standard methods for benchmarking vision algorithms (Deep Nets and others) are inadequate and can yield misleading impressions about their strengths (``an algorithm is only as good as the dataset it is evaluated on and the performance measures used'') and so tougher and more challenging performance measures should be developed. This is particularly important as these algorithms are applied to real world problems and hence have real world consequences.

The organization of this article is as follows. In Section~\ref{sec:history} we discuss the history of neural networks and its tendency to cycles of  boom and bust. Section~\ref{sec:success} describes a few of the successes of Deep Nets  while also mentioning some caveats. In Section~\ref{sec:understanding} we discuss the limited understanding of the internal workings of Deep Nets. Section~\ref{sec:cogsci} surveys their potential for helping to construct theories of biological visual systems, but also their limited relationships to real neurons and neural circuits. In Section~\ref{sec:challenges} we discuss the challenges that Deep Nets are now grappling with. Section~\ref{sec:explosion} is more speculative and argues that as vision researchers attempt to model increasingly complex visual tasks they will face a combinatorial explosion which Deep Nets may be unable to overcome and which may require computer vision researchers to revise the ways they evaluate vision algorithms. In Section~\ref{sec:outside} we briefly point out that the issues in this paper, in particular performance evaluations and dataset bias, matter outside academia and failure to address them properly risks leading to bad societal consequences. 

\section{Some History \label{sec:history}}

We are in the third wave of neural network approaches. The first two waves --- 1950s--1960s and 1980s--1990s --- generated considerable excitement but slowly ran out of steam. Despite a few exceptions, the overall performance of neural networks was disappointing for machines (artificial intelligence, machine learning) and for understanding biological vision systems (neuroscience, cognitive science, psychology). But the third wave --- 2000s--present --- stands out because of the dramatic success of Deep Nets on many large benchmark problems and their growing use for industrial application to real world tasks (e.g., face recognition). It should be acknowledged  that many of the currently successful neural networks, such as convolutional neural networks~\citep{DBLP:journals/neco/LeCunBDHHHJ89} and recurrent neural networks~\citep{rumelhart1986learning} were developed during the second wave. But their strengths were not appreciated until the availability of big datasets and the ubiquity of powerful computers (e.g., GPUs) which only became available after 2000 and which fueled the third wave.

The rise and fall of these neural network waves reflect changes in intellectual fashion and the varying popularity of other approaches. The second wave of neural networks was partly driven by the perceived limitations of classic artificial intelligence where disappointing results led to an AI winter in the mid-1980s (the first author observed this first hand as a researcher in the AI lab at MIT). In turn, the decline of the second wave corresponded to the rise of support vector machines, kernel methods, and related approaches. Credit is due to those neural network researchers who carried on despite discouragement through the troughs of the waves when it was sometimes hard to publish neural network papers. The pendulum has now swung again and it sometimes seems hard to publish anything that is not neural network related. We believe progress would be faster if researchers resisted the attraction of fashions and instead pursued a diversity of approaches and techniques. It is also unfortunate that students may not learn older techniques, like GrabCut~\citep{DBLP:journals/tog/RotherKB04}, Superpixels~\citep{DBLP:journals/pami/AchantaSSLFS12}, and Belief Propagation~\citep{DBLP:books/daglib/0066829}, and may not want to use them even if they do. For example, the first author is involved in a medical imaging project where a collaborator from the Medical School had to implement a superpixel algorithm because his own graduate students (in computer science) lacked the necessary expertise or motivation.

%But this relationship should be treated with caution, unless expert neuroscientists are involved. Real neurons are much more varied and complex than artificial neurons (there are fifty different types of neurons in the retina alone and also a range of different morphological structures) and important properties such as the neural code are only partially understood. In the short term, it may be best to think of artificial neural networks as a way of doing statistics (as researchers at early neural network meetings started speculating as we shared ski lifts), for example by interpreting Deep Nets as a sophisticated form of probabilistic regression, instead of as models of real neurons. ``Neuronal plausibility'' is a desirable property of a cognitive system but it is not always easy to pin down. The relationships between neural networks to real neural circuits is a very challenging research problem that may take years to resolve. Similarly there is a big gap between human cognition and what current neural networks can perform. Clearly neural networks, and other big data methods, are very important techniques for studying the brain and the mind but there remain huge gaps between these disciplines.

\section{The Initial Successes, With Some Caveats}
\label{sec:success}

\newcommand{\rt}{figs}
\newcommand{\wdtg}{.485\linewidth}
\newcommand{\ft}[1]{#1}
\newcommand{\resfigwdt}[1]
{
	{\setlength{\tabcolsep}{.2em}	
		%\begin{tabular}{c@{}c@{}c@{}c@{}c@{}c@{}c@{}c}
		\begin{tabular}{cc}
			\ft{Input} & \ft{Boundaries} \\
			\includegraphics[width=\wdtg]{\rt/l#1_input}&
			\includegraphics[width=\wdtg]{\rt/l#1_boundaries}\\
			 \ft{Surface Normals} & \ft{Saliency}\\	
			\includegraphics[width=\wdtg]{\rt/l#1_surface_normals.png}&
			\includegraphics[width=\wdtg]{\rt/l#1_saliency}\\
			 \ft{Semantic Segmentation} & \ft{Semantic Boundaries}\\
			\includegraphics[width=\wdtg]{\rt/l#1_semantic_segmentation.png}&  
			\includegraphics[width=\wdtg]{\rt/l#1_semantic_boundaries.png}\\
			 \ft{Human Parts} &  \ft{Detection} \\
			\includegraphics[width=\wdtg]{\rt/l#1_human_parts.png} & 
			\includegraphics[width=\wdtg]{\rt/l#1_detection}
		\end{tabular}
	}
}
\begin{figure}[t]
\resfigwdt{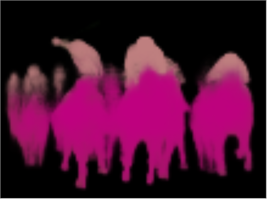}
\caption{A wide variety of vision tasks can be performed by Deep Nets. These include: boundary detection, semantic segmentation, semantic boundaries, surface normals, saliency, human parts, and object detection. Figure taken from \citet{DBLP:conf/cvpr/Kokkinos17}, which illustrated one model for multi-task learning. }
\label{fig:ubernet}
\end{figure}

The computer vision community was skeptical about Deep Nets until the impressive performance of AlexNet  \citep{DBLP:conf/nips/KrizhevskySH12} for classifying objects in ImageNet \citep{DBLP:conf/cvpr/DengDSLL009}\footnote{The first author remembers that in the mid 1990s and early 2000s the term ``neural network'' in the title of a  submission to a computer vision conference was sadly a good predictor for rejection and recalls sympathizing with researchers who were pursuing such unfashionable ideas.}. This classification task assumes there is a foreground object which is surrounded by a limited background region, so the input is similar to one of the red boxes of the bottom right image in Figure~\ref{fig:ubernet}. AlexNet's success stimulated the vision community leading to a variety of Deep Net architectures with increasingly better performance on object classification, first designed by human experts \citep{DBLP:journals/corr/SimonyanZ14a, DBLP:conf/cvpr/HeZRS16}, then assisted by machines \citep{DBLP:conf/iclr/ZophL17, DBLP:conf/eccv/LiuZNSHLFYHM18}. We will elaborate on the latter approach --- known as Neural Architecture Search --- in Section~\ref{sec:nas}.

Deep Nets were also rapidly adapted to other visual tasks such as object detection, where the image contains one or more objects and the background is much larger, e.g., the PASCAL challenge \citep{DBLP:journals/ijcv/EveringhamGWWZ10}. For this task, Deep Nets were augmented by an initial stage which made proposals for possible positions and sizes of the objects and then applied Deep Nets to classify the proposals~\citep{DBLP:conf/cvpr/GirshickDDM14} (current methods train the proposals and objects together in what is called ``end-to-end''). These methods outperformed the previous best methods, the Deformable Part Models \citep{DBLP:journals/pami/FelzenszwalbGMR10}, for the PASCAL object detection challenge (PASCAL was the main object detection and classification challenge before ImageNet). Other Deep Net architectures also gave enormous performance jumps in other classic tasks like edge detection~\citep{DBLP:conf/iccv/XieT15}, semantic segmentation~\citep{DBLP:conf/cvpr/LongSD15, DBLP:journals/pami/ChenPKMY18}, occlusion detection~\citep{DBLP:conf/eccv/WangY16} (edge detection with border-ownership), symmetry axis detection~\citep{DBLP:journals/tip/ShenZJWBY17}. Major increases also happened for human joint detection~\citep{DBLP:conf/nips/ChenY14, DBLP:conf/cvpr/ChenY15}, human part segmentation~\citep{DBLP:conf/eccv/XiaWCY16}, binocular stereo~\citep{DBLP:conf/cvpr/ZbontarL15, DBLP:conf/cvpr/MayerIHFCDB16}, 3D depth estimation from single images~\citep{DBLP:conf/nips/EigenPF14}, and scene classification~\citep{DBLP:conf/nips/ZhouLXTO14}. Several of these tasks are illustrated in Figure~\ref{fig:ubernet}.

But although Deep Nets are very effective, almost always outperforming alternative techniques when evaluated on benchmark datasets, they are not general purpose and their successes come with the following three restrictions. 

Firstly, like almost all machine learning algorithms Deep Nets are designed for specific visual tasks. Most are designed for single tasks and a Deep Net designed for one task will not be well-suited for another. For example, a Deep Net designed for object classification on ImageNet cannot perform human parsing (i.e. the detection of human joints) on the Leeds Sports Dataset (LSD). There are, however, some exceptions and \emph{transfer learning}  sometimes makes it possible to adapt Deep Nets trained on one task to a closely related task provided annotated data is available for that task (see Section~\ref{sec:transfer}). Intuitively this happens because the features learned by the Deep Net captures image structures that are useful for both tasks. In addition, researchers have recently developed Deep Nets which can perform multiple tasks. UberNet~\citep{DBLP:conf/cvpr/Kokkinos17} is an example which also gives an introduction to the literature. But, in general, there has been a growing zoo of different Deep Net architectures designed for specific tasks which include cascades of networks and supervision at several different levels of the network although the recent introduction of Neural Architecture Search, see Section~\ref{sec:nas}, offers the promise of a more principled way of exploring this zoo and perhaps even discovering universal Deep Net features.

\begin{figure}[t]
\centering
\includegraphics[width=\linewidth]{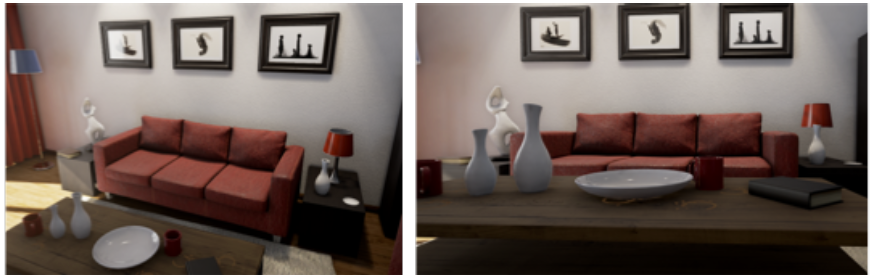}
\includegraphics[width=\linewidth]{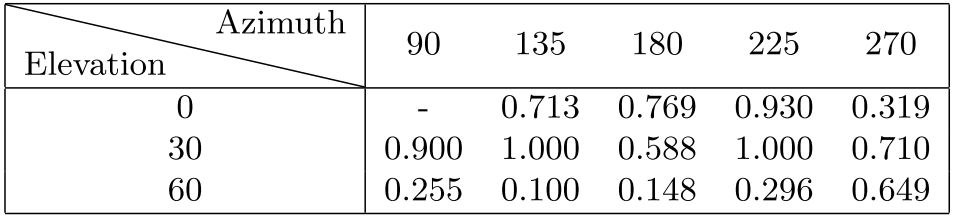}
\caption{Figure taken from \citet{DBLP:conf/eccv/QiuY16}. UnrealCV allows vision researchers to easily manipulate synthetic scenes, e.g. by changing the viewpoint of the sofa. We found that the Average Precision (AP) of Faster-RCNN \citep{DBLP:conf/nips/RenHGS15} detection of the sofa varies from 0.1 to 1.0, showing extreme sensitivity to viewpoint. This is perhaps because the biases in the training cause Faster-RCNN to favor specific viewpoints.}
\label{fig:unrealcv}
\end{figure}

Secondly, Deep Nets which perform well on benchmark datasets may fail badly on real world images outside the dataset. A Deep Net is only as good as the dataset it has been evaluated on (and the performance measures used). This is because the set of real world images is infinitely large and so it is hard for any dataset, no matter how big, to be representative of the complexity of the real world. This is an important issue which we will return to in Section~\ref{sec:explosion}. For now, we simply remark that all datasets have biases. These biases were particularly blatant in the early vision datasets and researchers rapidly learned to take advantage of them for example by exploiting the background context  (e.g., detecting fish in Caltech101 was easy because they were the only objects whose backgrounds were water). Comparative studies showed that methods which performed well on some datasets often failed to generalize to others \citep{DBLP:conf/cvpr/TorralbaE11}. These problems are reduced, but still remain, despite the use of big datasets and Deep Nets. For example, background context remains problematic even for ImageNet as shown by work where the target object is masked out but can be predicted with reasonable accuracy from the context \citep{DBLP:conf/ijcai/ZhuXY17}. Biases also occur if the dataset contain objects from limited viewing conditions, e.g., as shown in Figure~\ref{fig:unrealcv}, a Deep Net trained to detect sofas on ImageNet can fail to detect them if shown from viewpoints which were underrepresented in the training dataset. In particular, Deep Nets are biased against ``rare events'' which occur infrequently in the datasets.  But in real world applications, these biases are particularly problematic since they may correspond to situations where failures of a vision system can lead to terrible consequences, e.g., datasets used to train autonomous vehicles almost never contain babies sitting in the road. Similarly, datasets may under-represent (and certainly fail to annotate) the hazardous factors which are known to cause algorithm to fail, such as specularity for binocular stereo. We will return to this example in Section~\ref{sec:sensitivity}.
% Although some of these problems can be reduced  by re-training Deep Nets on additional data \chenxi{This sentence is incomplete} We will return to this issue later \todo{forward references to later sections}. 

Thirdly, almost all Deep Nets require annotated data for training and testing. This has the effect of biasing vision researchers to work on visual tasks for which annotation is easy instead of on problems which are important. For example, annotation for object detection merely requires specifying a tight bounding box around an object which is fairly straightforward to do. But for other vision tasks, such as detecting the joint of a human of per-pixel annotation of objects, annotation is much harder and for some tasks such as estimating 3D depth it is almost impossible. There are methods which reduce the need for supervision as discussed in Section~\ref{sec:transfer}, and there is also the possibility of using synthetic stimuli (generated by computer graphics engines) which enables groundtruth to be available for all visual tasks. But realistic synthetic stimuli are limited and so the computer vision community is only gradually, and somewhat reluctantly, starting to use it.

In summary, Deep Nets are a set of tools which are constantly being refined and developed according to the needs of specific visual tasks. The greatest successes of Deep Nets have relied heavily on fully supervised data, though there are growing advances using less supervision which we will discuss in Section~\ref{sec:challenges}.
Their performance can fail to generalize to images outside the dataset they have been trained on and, as we will discuss in Section~\ref{sec:explosion}, this is particularly problematic due to  the infinite complexity of real world images.

\section{Towards Understanding Deep Nets}
\label{sec:understanding}

%\begin{figure*}[t]
%\centering
%\includegraphics[width=\linewidth]{figs/cls_vis_other_small}
%\caption{Figure taken from \citet{DBLP:journals/corr/WangZPY15}. The visual concepts obtained by population encoding are visually tight and we can identify the parent %object class pretty easily by just looking at the mid-level concepts.}
%\label{fig:vc}
%\end{figure*}

It is  difficult to characterize what Deep Nets can do and to understanding their inner workings. Theoretical results show that multi-layer perceptrons, and hence Deep Nets, can represent any input output function provided there are a sufficient number of hidden units \citep{DBLP:journals/mcss/Cybenko89, DBLP:journals/nn/HornikSW89}. But, as the first author can testify from personal experience   \citep{DBLP:journals/nn/XuKY94}, theoretical results which hold in the asymptotic limit are of limited utility. Much more valuable would be results which hold for limited numbers of hidden units and limited training data, but it is hard to see what meaningful theoretical results could be obtained for systems as complicated as Deep Nets.

At a more intuitive level it seems possible to get some rough understanding of Deep Nets at least when applied to visual tasks. The hierarchical structure of Deep Nets is similar to classical models of the visual cortex such as the NeoCognition \citep{fukushima1982neocognitron} and HMax \citep{riesenhuber1999hierarchical} and captures many of the intuitions which motivated these models. Deep Nets contain feature representations where those at lower levels have receptive fields of limited sizes and which are sensitive to the precise positions of patterns. But as we ascend the hierarchy the receptive fields become larger and  more sensitive to specific patterns, while being less concerned about their exact locations. 

This can be partially understood by studying the  activities of the internal filters/features of the convolutional levels of Deep Nets \citep{DBLP:conf/eccv/ZeilerF14, DBLP:journals/corr/YosinskiCNFL15}. In particular, if Deep Nets are trained for scene classification then some convolutional layer filters roughly correspond to objects which appear frequently in the scene, while if the Deep Nets are trained for object detection, then some features roughly correspond to parts of the objects \citep{DBLP:journals/corr/ZhouKLOT14}. Detailed studies of feature properties for a restricted class of objects (e.g., vehicles) and with fixed object scales show that clusters of feature responses are often interpretable and correspond to subparts of the objects \citep{DBLP:journals/corr/WangZPY15}.\footnote{In addition to visualization, training a small neural network (also known as readout functions) on top of deep features is another popular technique of assessing how much they encode some particular properties, which is now widely adopted in the self-supervised learning literature \citep{DBLP:conf/eccv/NorooziF16, DBLP:conf/eccv/ZhangIE16}.} But we acknowledge that while these studies are encouraging they remain mostly impressionistic and lack the precision of true understanding (e.g., these studies have not yet enabled researchers to learn models of objects and object-parts in an unsupervised manner).

This suggests the following rough conceptual picture of Deep Nets. The convolutional levels represent the manifold of intensity patterns at different levels of abstraction. The lowest levels represent local image patterns while the high levels represent larger patterns which are invariant to the details of the intensity patterns. From a related perspective, the weight vectors represent a dictionary of templates of image patterns. The final ``decision layers'' of the Deep Net are usually harder to interpret  but it is plausible that they make decisions based on the templates represented by the lower layers.\footnote{Admittedly in ResNets \citep{DBLP:conf/cvpr/HeZRS16}, there is only one ``decision layer'', and the analogy to ``template matching'' also weakens at higher layers due to presence of the residual connection. } This ``dictionary of templates'' interpretation of Deep Nets suggests they are very efficient to learn and represent an enormous variety of image patterns, and interpolate between them, but cannot extrapolate much beyond the patterns they have seen in their training dataset. Other studies suggest that Deep Nets are less effective at modeling visual properties which  are specified purely by geometry, particularly if the input consists of binary valued patterns corresponding to the presence or absence of boundary edges. It is an open issue whether Deep Nets can learn features that ``factorize'' different visual properties which, as we will argue later in Section~\ref{sec:explosion}, will ultimately be necessary for dealing with the full complexity of real images.

% Hence, despite Deep Nets  achievements on MNIST, they may be less suited to reading compared to other approaches which factorize geometry and appearance (see later section).
\section{Deep Nets and Biological Vision}
\label{sec:cogsci}

Deep Nets have much to offer for studying biological vision systems and, in particular, disciplines like cognitive science, neuroscience and psychology which aim at understanding the mind and the brain. They can help develop and test computational theories by  exploiting the availability of big data while raising the possibility of understanding the  brain by relating the artificial neurons in Deep Nets to real neurons in the brain. In turn, studies of biological vision show that human infants acquire visual abilities in ways that are very different from current machine learning algorithms, which may suggest alternative learning strategies. In addition, real biological neurons and neural circuits are very different from those in artificial neural networks, which suggests new neural architectures to explore.

\subsection{Exploiting Big Data}

The use of Deep Nets, and other machine learning techniques, can help develop theories of mind and brain which exploit big data. This can be done in roughly three ways. Firstly, Deep Nets can motivate biological vision researchers to extend their theories beyond simplified stimuli and deal with the enormous complexity of real world images. Secondly, they can be used to partially learn the  knowledge about the visual world that humans and other animals obtain through development and experience. Thirdly, they enable theories to be tested on complex stimuli and compared to alternative theories. We will now address these issues in turn.

Historically, studies of biological visual systems have largely relied on simple synthesized stimuli, such as Julesz random dot stereograms~\citep{julesz1971foundations}, or sinusoidal gratings, or Gabor functions: see \citep[Chapter~12]{arbib2016neuron} for more background on these examples. These studies have led to many important findings and were historically necessary because the complexity of natural image stimuli means that it is extremely hard to perform controlled scientific experiments by systematically varying the experimental parameters. This also follows the well established scientific strategy of divide and conquer which aims at understanding by breaking down complex phenomena into more easily understandable chunks. But studying vision on simplified stimuli has limitations which Deep Nets and big data can help address. As researchers in computer vision discovered in the 1980s, findings on simplified synthetic stimuli such as blocks world \citep{guzman1968decomposition} --- though sometimes providing motivations and good starting points --- typically required enormous modifications before they could be extended to realistic stimuli if they could be extended at all. Computer vision researchers had to leave their comfort zone of synthetic stimuli and address the fundamental challenge of vision: namely how  visual systems deal with the complexity and ambiguity of real world images and achieve the superpower \citep{changizi2010vision} of converting the light rays that enter the eye, or a camera, into an interpretation (both geometric and semantic) of the three-dimensional physical world. Driven by the need to address these issues, computer vision researchers developed a large set of mathematical and computational techniques and increasingly realized the importance of learning theories from data using tools like Deep Nets, which required large annotated datasets. The same techniques can be directly applied to studying biological vision by predicting experimental responses to visual stimuli, e.g.,  human performance in behavioral experiments,  the responses of neurons, or fMRI activity.

Big data, and learning methods for mining the data, are particularly important for vision because, as  leading vision scientists like Gregory \citep{gregory1973eye}, Gibson \citep{gibson1986ecological} and Marr \citep{marr1982vision} have argued, visual systems require knowledge of the world in the form of natural and ecological constraints. In Gregory's words ``perception is not just a passive acceptance of stimuli, but an active process involving memory and other internal processes''. In other words, the visual systems of humans, and other animals, exploit a large amount of  knowledge which has been acquired through development and experience  \citep{arterberry2016development}. Big data methods, like Deep Nets, gives a surrogate way for vision scientists to partially learn this knowledge by studying properties of real world images. 

Finally, the use of big datasets are also very important for testing visual theories because they enabled detailed comparisons with alternative theories. They make it easy to  reject ``toy theories'' that exploit the biases inherent in small datasets and simplified stimuli. In summary, the use of Deep Nets and big data  enable biological vision researchers to develop and test theories that can work in realistic visual domains and address the fundamental challenge of vision. 

\subsection{Real Neurons and Neural Circuits}

From the neuroscience perspective, Deep Nets have been used to predict brain activity, such as fMRI and other non-invasive measurements, and there are a growing number of examples \citep{cichy2016comparison, wen2017neural}. They have also been applied to predicting neural responses as measured by electrophysiology such as  predicting the response of neurons in the ventral stream \citep{yamins2014performance}. These are examples where Deep Nets' ability to learn from data and to deal with the complexity of real stimuli really pays off. But in terms of understanding the neuroscience of the primate ventral stream,  this is best thought of as a starting point. The ventral stream of primates is very complex and  there is evidence that it estimates the three-dimensional structure of objects and parts \citep{yamane2008neural}, possibly relating  to the  classic theory of object recognition by component \citep{biederman1987recognition}. More generally, the  primate visual systems must perform all the visual tasks listed in Section~\ref{sec:success}, namely edge detection, binocular stereo, semantic segmentation, object classification, scene classification, and 3D depth estimation. The vision community has developed a range of different Deep Nets for these tasks so it is extremely unlikely, for example, that  a Deep Net trained for object classification on ImageNet would be able to account for the richness of the primate visual systems though, as discussed earlier, the low-level features may be similar for networks which perform different visual tasks.  

It should also be emphasized that while Deep Nets perform computations bottom-up in a feedforward manner there is considerable evidence of top-down processing in the brain \citep{lee2003hierarchical}, particularly driven by top-down attention \citep{gregoriou2014lesions}. Researchers have also identified cortical circuits \citep{mcmanus2011adaptive} which implement spatial interactions (though possibly in a bottom-up and top-down manner). These types of phenomena require other families of mathematical models, perhaps the compositional models described in Section~\ref{sec:explosion}.

But, more fundamentally, it must be acknowledged that there are big differences between the artificial neurons used in Deep Nets and real neurons in the brain. Artificial models of neurons are, at best, great simplifications of realistic neurons as shown by studies of real neurons in vitro \citep{poirazi2001impact}. Neuroscientists have found that there are over one hundred different types of neurons, and there are enormous morphological differences which may be exploited to enable computation \citep{seung2012connectome}. There is also lack of detailed understanding of neural circuits.  For example, the wiring diagram of C-elegans has been known for over thirty years but there is still only limited understanding of how it functions as a neural circuit. As stated by O. Hobert the wiring diagram ``is like a road map that tells you where cars can drive, but does not tell you when or where cars are actually driving'' \citep{jabr2012connectome}.  Understanding real neurons and real neural circuits is a fascinating scientific challenge and exciting new engineering techniques, e.g.,  \citep{boyden2005millisecond}, combined with the availability of huge datasets and the tools to analyze them suggest that we may be close to gaining insight into biological neural networks that can be used to develop more advanced artificial neural networks.

\subsection{Cognitive Abilities: Deep Nets and Scientific Understanding}

Deep Nets, and other machine learning techniques, are very helpful tools for modeling human visual cognition. But the human visual system is, in general, much superior to  current state-of-the-art computer vision algorithms and outperforms them in multiple dimensions. Firstly, as discussed in the introduction, it performs a multitude of visual tasks at the same time  (e.g., detect objects, parse them into parts, find their boundaries, and estimate their 3D configurations). Secondly, humans can learn very efficiently from small numbers of examples presumably by exploiting prior knowledge and physical properties of the world. Thirdly, the human visual system is robust to viewpoint changes, to novel contexts, to partial occlusion (including overlapping objects as in CAPTCHAs as will be discussed in Section~\ref{sec:explosion}), and to pixel-level adversarial image attacks. Fourthly, humans can learn on one image domain, e.g., real images, line drawing, or visual arts, and effortlessly transfer to new domains (often with no supervision). In general, studies of cognitive science show that human visual systems can work at levels of abstraction which current Deep Nets cannot match. This can be illustrated by human's ability at visual analogies: some of which depend only on visual similarity but others depend on the notion of parts and subparts, while others include the idea of function. From another perspective, it can also be argued that the goal of vision science is to discover underlying principles. From this perspective, a model that explains phenomena in terms of an uninterpretable Deep Net would not be very satisfying. This is a debatable issue on which reasonable people can disagree. But we suspect that progress in AI will also require interpretable models partly for the pragmatic engineering principle, that this is necessary for debugging and for performance and safety guarantees. %A possible compromise is that Deep Nets would serve to perform the signal-to-symbol task of extracting symbolic representations from images

Studies also suggest that humans learn vision in ways that are very different from current machine learning approaches. Human infants learn vision without direct supervision. There is  an enormous literature on how infants learn vision and different visual abilities arise at different times in a stereotyped sequence \citep{arterberry2016development}. Infants learn by actively interacting with and exploring the world and are not merely passive acceptors of stimuli \citep{gregory1973eye}. They are more like tiny scientists who understand the world by performing experiments and seeking causal explanations for phenomena \citep{gopnik1999scientist}.

Some researchers have argued that Deep Nets have superhuman powers. But these claims rarely survive careful inspection and can be due to Deep Nets fitting the biases in the datasets on which the studies are performed, as discussed earlier~\citep{DBLP:conf/ijcai/ZhuXY17}. The few exceptions where Deep Nets can potentially outperform humans are in situations for which evolution and experience put humans at a disadvantage. For example, AI systems can outperform humans by recognizing hundreds of millions of faces provided they are seen from front-on under reasonable lighting conditions and with limited occlusion, but until recently most humans never saw more than a few thousand people in their whole lifetime. There are, however, serious concerns that the results of these AI systems can be affected by dataset bias causing problems for some ethnic minorities (see Section~\ref{sec:outside}). It is also possible that AI systems could perform better than the average radiologists when reading computer tomography (CT) images, because  even the most expert radiologists have only seen a fairly small number of CT scans and AI systems can directly access the  three-dimensional data in CT scans, while radiologists can only view two-dimensional slices. In each of these cases, humans are at a disadvantage because they do not have access to, and hence cannot exploit, the enormous amounts of annotated big data which enable Deep Nets to do so well on these tasks. It should be mentioned that the human visual system suffers from visual illusions, but these are often for impoverished stimuli where there are several possible interpretations and humans often appear to be performing a sensible strategy.\footnote{\url{https://michaelbach.de/ot/}} Other examples where humans make errors can be considered as accidental alignments, for example a woman standing on a beach towel appears to be on a flying carpet because of a shadow that appears to be cast by the towel but is really cast by a flagpole outside the image \citep[Chapter~12]{arbib2016neuron}. There are other aspects of the human visual system such as change blindness \citep{rensink1997see} and failure to see a gorilla \citep{simons1999gorillas} which computer vision systems would not want to mimic and should be able to avoid. 

Recent work \citep{DBLP:journals/pnas/Firestone20} proposes studying the differences between human and machine perception using Chomsky's distinction between performance and competence \citep{chomsky2014aspects}. Are comparisons between Deep Nets and humans fair and do they take into account the different underlying capabilities of human and machine computational resources? Can we distinguish between visual tasks that humans can do and deep networks are incapable of without taking the competence versus performance distinction into account? There may be certain tasks that deep networks are fundamentally unable to do, similar to those tasks which distinguish human cognitive abilities from those of other animals \citep{penn2008darwin}, but can we identify them without thinking carefully about these issues?  Perhaps deep networks trained with enough annotations, better loss functions, and more data would be sufficient to overcome some, or perhaps even all, of their current limitations with respect to human vision? Our opinion is that this is unlikely, without requiring modifications to the algorithms which are so significant that they should probably be renamed, but we cannot say for sure. Nevertheless \citep{DBLP:journals/pnas/Firestone20}  quotes examples showing there are a lot of similarities between deep networks and humans, e.g., despite deep network's vulnerability to adversarial attacks it can be shown that the ``human subjects correctly anticipated the machine’s preferred label over relevant foils'' \citep{zhou2019humans}.

From the perspective of the study of human perception it is, however, not surprising that there are situations where machines can outperform humans. Ideal observer theory, see  \citet{geisler2011contributions, tjan1995human, liu1995object}, is a classical perceptual science technique  \citep{GreenAndSwets1966} which compares performance of human observers against an ideal observer for specific visual tasks. The ideal observer is assumed to know the probability models that generate the data and hence is able to make the optimal bayesian prediction. Not surprisingly, humans generally do much worse than the ideal observers, sometimes by orders of magnitude. For example, in a study of motion perception, \citet{barlow1997correspondence} showed that human performance was badly degraded compared to an ideal observer model where approximations were made to simply computation. In follow up work, \citet{lu2006ideal} showed that if the true ideal observer was calculated then it outperformed human observers by orders of magnitude not only on the original experiments reported in \citet{barlow1997correspondence} but also on additional related experiments. \citet{lu2006ideal} also showed that human performance was much better predicted by a generic motion model which, unlike an ideal observer, had no knowledge of the experimental stimuli. In general, ideal observer analysis has always focused on highly constrained tasks where the generative process is narrowly specified so that the optimal bayesian prediction can be found by the ideal observer, and the human observer has the same well-defined task. But the task does not necessarily match what humans are good at doing. Perhaps, more generally, both  deep nets and ideal observers can do better than humans not only because they can exploit properties of the stimulus humans cannot (because the stimuli are a subset of the much bigger set of stimuli that can occur in the real world) but also because they are doing a task that may have low priority for humans. A computer vision app can identify the hundreds of plants in the first author's garden much better than the first author can, but a plant expert can surely outperform it.

In summary,  Deep Nets, and other techniques which exploit big data, are  a tool that mind and brain researchers should know how to use and not misuse. At present, Deep  Nets fail to capture some of the most interesting phenomena such as human's ability to perform abstractions and perform analogical reasoning (although Deep Nets might be useful as building blocks to construct such a theory). Nevertheless a closer relationship between biological and artificial models of vision would be beneficial to both disciplines. Researchers in AI have developed a large set of technical tools,  like Deep Nets, which can allow their models to be applied to the complexity of natural images and tested under rigorous realistic conditions. Vision scientists can challenge computer vision researchers to develop algorithms which can perform as well as, or better than humans, in challenging situations while using orders of magnitude less power than current computers.

%It can also be questioned whether Deep Nets are really satisfactory as theories of Cognitive Science or Neuroscience. A common criticism is that they are merely black boxes and hence do not capture or explain internal representations or other cognitive processes. This echoes a similar criticism of their use for Artificial Intelligence applications. But, as with AI, this may only be a temporary limitation and that better understanding of Deep Nets, perhaps along the lines of Section~\ref{sec:understanding}, and the development of more interpretable, but equally effective, models will help alleviate it. Similar views were expressed in the second wave of neural networks \citep{mccloskey1991networks}. It remains possible that the best theory of the brain, or any really complex physical system, may only be a black box. But this seems too pessimistic and we would be very surprised if the long-term solution to AI or Neuroscience is an uninterpretable black box (partly due to scaling and diagnostic issues which will be discuss in Section~\ref{sec:explosion}).

\section{The Frontiers and Challenges}
\label{sec:challenges}

This section describes some of the current frontiers and challenges of Deep Nets and the attempts to address them. Some of these challenges are gradually being overcome while others, such the sensitivity to non-local attacks, may require more fundamental changes as we speculate in Section~\ref{sec:explosion}.

\subsection{Relaxing the Need for Full Supervision}
\label{sec:transfer}

A disadvantage of Deep Nets is that they typically need a very large amount of annotated training data, which restricts their use to situations where big data is available. But this is not always the case. In particular, transfer learning shows that the features of Deep Nets learned on annotated datasets for certain visual tasks can sometimes be transferred to novel datasets and related tasks, thereby enabling learning with much less data and sometimes with less supervision. For example, as mentioned earlier, Deep Nets were first successful for object classification on ImageNet but had previously failed on object detection on the smaller PASCAL dataset. This was presumably because PASCAL was not sufficiently large for training a Deep Net, while ImageNet was almost two orders of magnitude larger than PASCAL. But researchers quickly realized that it was possible to train a Deep Net for object detection and semantic segmentation on PASCAL by initializing the weights of the Deep Net by the weights of a Deep Net trained on ImageNet \citep{DBLP:conf/cvpr/GirshickDDM14, DBLP:conf/cvpr/LongSD15, DBLP:journals/pami/ChenPKMY18}. This also introduced a mechanism for generating proposals, see Figure~\ref{fig:ubernet} (bottom right). It wasn't until a few years later when researchers developed specialized architectures to train on PASCAL without the need for pre-training \citep{DBLP:conf/cvpr/JegouDVRB17, DBLP:conf/iccv/ShenLLJCX17}.

% More generally, researchers found that they could transfer features from Deep Nets trained on one task on one dataset to perform related tasks on a second dataset. In some cases, this consisted of simply using the first dataset to initialize the weights when training on the second (so that the final values of the weights, particularly for the higher levels, may have little to do with their initial values) while in other situations, the weights changed little and were similar for both tasks and/or datasets.  \todo{Rewrite these two paragraphs and make the distinction between pretraining as initialization and finetuning. Wei --- DeepLab --- Feng. }\chenxi{This seems like the exact same point as the previous paragraph.} For example, researchers showed that Deep Nets trained for face verification could be transferred to the related task of facial pain estimation \citep{DBLP:conf/icip/WangXLTRHQCY17}. This is presumably because the two tasks required fairly similar image representations capturing the fine-scale appearance of facial features.

This ability to transfer Deep Net knowledge learned on another domain relates intuitively to the way children learn. A child initially learns rather slowly compared to other young animals but at critical periods the child's learning accelerates very rapidly \citep{smith2005development}. From the ``dictionary of templates'' perspective, this could happen because after a child has learned to recognize enough objects he/she may have enough building blocks (i.e. deep network filters) to be able to represent new objects in terms of a dictionary of existing templates. If so, only a few examples of the new object may be needed in order to do few-shot learning.

Few-shot learning of novel object categories has been shown for Deep Nets provided they have first been trained on a large set of object categories \citep{DBLP:conf/iccv/MaoWYWHY15, DBLP:conf/nips/VinyalsBLKW16, DBLP:conf/cvpr/QiaoLSY18}. Another strategy is to train a Deep Net (technically a \emph{Siamese network}) to learn similarity on the set of object categories, hence obtaining a similarity measure for the new objects. For example, \citet{DBLP:journals/corr/LinWLZYL17} trained a Siamese network to learn similarity for objects in ShapeNet \citep{DBLP:journals/corr/ChangFGHHLSSSSX15} and then this similarity measure was used to cluster objects in the Tufa dataset \citep{DBLP:journals/jmlr/SalakhutdinovTT12}. Other few-shot learning tasks can also be done by using features from Deep Nets trained for some other tasks as ways to model the visual patterns of objects.

More recently, there has been work on unsupervised learning which shows that optical flow and structure from motion can be learned without requiring detailed supervision but only an energy function model \citep{DBLP:conf/aaai/RenYNLYZ17, DBLP:conf/cvpr/ZhouBSL17}. Like many neural nets in the third wave some of the  ideas can be found in obscure papers from the second wave \citep{smirnakis1995neural}. In some cases, this can even be bootstrapped to learning depth from single images. Other forms of unsupervised learning show that Deep Net features can be learned by tracking an object over time \citep{DBLP:conf/iccv/WangG15}, or by distinguishing between scrambled and unscrambled images \citep{DBLP:conf/iccv/DoerschGE15, DBLP:conf/eccv/NorooziF16}, or by contrastive learning which distinguishes between parts or views of the same image and those of a different image \citep{DBLP:conf/cvpr/WuXYL18, DBLP:journals/corr/abs-1911-05722}. Recently this last line of research has received considerable attention in the computer vision community.

Other studies show that Deep Nets can exploit large numbers of unsupervised, or weakly supervised, data provided they have sufficient annotated data to start with. For example, to train object detection using images where only the names of the objects in the image are known but their locations and sizes are unknown. This is known as weakly supervised learning and it can be treated as missing/hidden data problem which can be addressed by methods such as Multiple Instance Learning (MIL) or Expectation-Maximization (EM). Performance of these types of methods is often improved by using a small amount of fully supervised training data which helps  the EM or MIL algorithms converge to good solutions, e.g., see \citet{DBLP:conf/iccv/PapandreouCMY15}.

\subsection{Defending Against Adversarial Examples}
\label{sec:adversarial}

\begin{figure}[t]
\centering
\includegraphics[width=\linewidth]{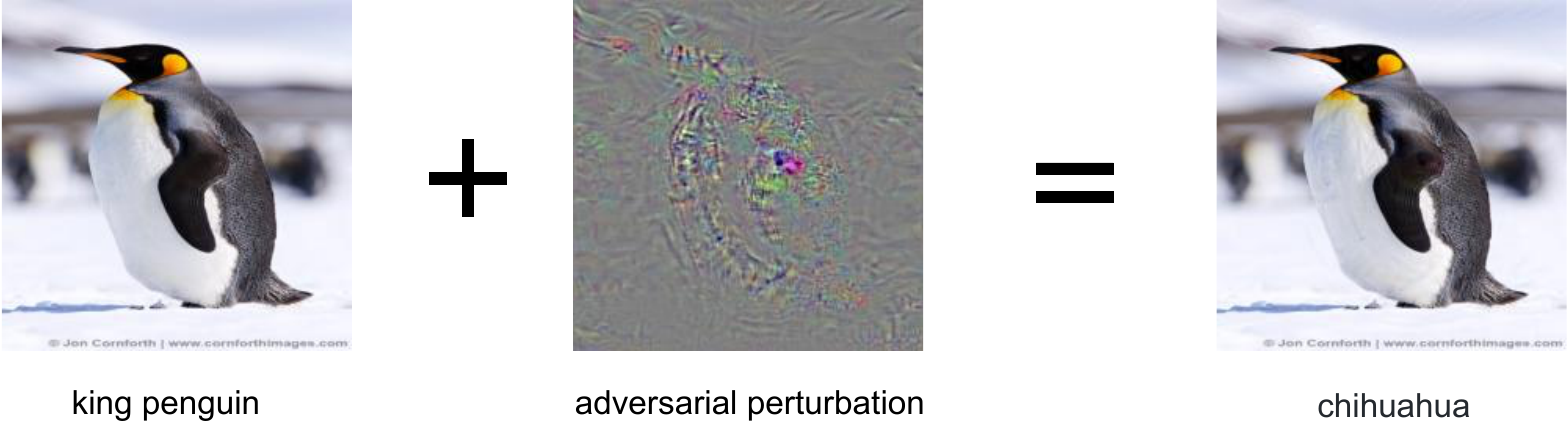}
\caption{Figure taken from \citet{DBLP:journals/corr/abs-1711-01991}. A deep network can correctly classify the left image as \textit{king penguin}. The middle image is the adversarial noise magnified by 10 and shifted by 128, and on the right is the adversarial example misclassified as \textit{chihuahua}.}
\label{fig:adv-cls}
\end{figure}

\begin{figure}[t]
\centering
\includegraphics[width=\linewidth]{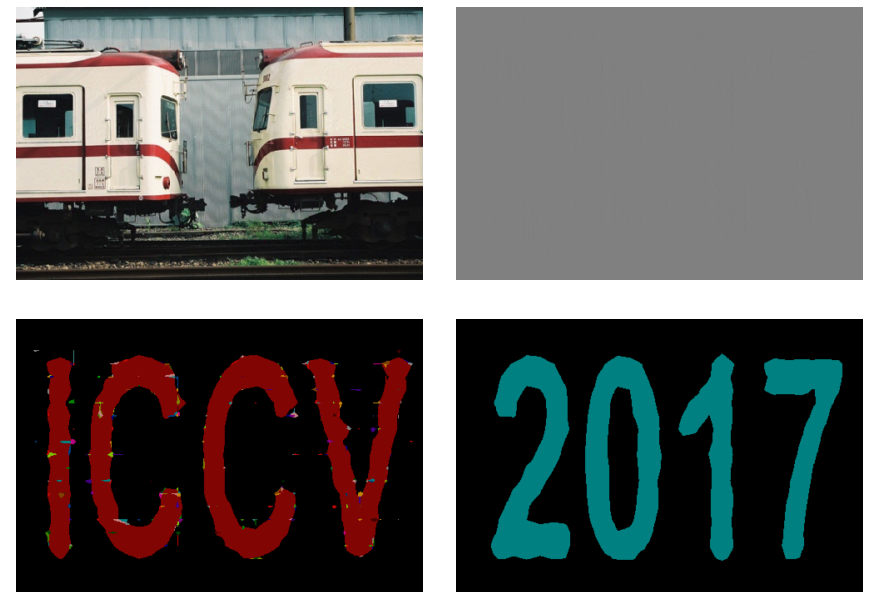}
\caption{Figure taken from \citet{DBLP:conf/iccv/XieWZZXY17}. The top row is the input (adversarial perturbation already added) to the segmentation network, and the bottom row is the output. The red, blue and black regions are predicted as \textit{airplane}, \textit{bus} and \textit{background}, respectively.}
\label{fig:adv-seg}
\end{figure}

Another limitation of Deep Nets comes from studies showing they can be successfully attacked by imperceptible modifications of the images which nevertheless cause the Deep Nets to make major mistakes for object classification \citep{DBLP:journals/corr/SzegedyZSBEGF13}, object detection, and semantic segmentation \citep{DBLP:conf/iccv/XieWZZXY17} (see Figure~\ref{fig:adv-cls} and Figure~\ref{fig:adv-seg}). This problem partly arises because the datasets are finite and contain only an infinitesimal fraction of all possible images. Hence there are infinitely many images arbitrarily close to the training images and so there is a reasonable chance that the Deep Net will misclassify some of them. Researchers have shown that they can find such images either by {\it white box} attacks, where the details of the Deep Net are known, or by {\it black box} attacks, when they are not. But there are now strategies which defend against these attacks. 
One idea is to introduce small random perturbations into the images \citep{DBLP:journals/corr/abs-1711-01991}, exploiting the assumption that the ``attack images'' are very unstable so small random perturbation will defend against them. 
Admittedly, \citet{DBLP:conf/icml/AthalyeC018, DBLP:conf/icml/UesatoOKO18} have shown that defenses of this nature can be circumvented if the attacker is aware that such defenses might be present.
Another strategy is to treat these ``attack images'' as extra training data, known as ``adversarial training'' \citep{DBLP:journals/corr/GoodfellowSS14, DBLP:journals/corr/MadryMSTV17}. This studies show that adversarial images should be considered a ``feature not a bug'' because they act as a sophisticated form of data augmentation which can result in more robust Deep Net features which are also more interpretable \citep{tsipras2019robustness}. It should also be acknowledged that adversarial attacks can be mounted against any vision algorithm and it would be much easier to successfully attack most other vision algorithms~\citep{DBLP:conf/pkdd/BiggioCMNSLGR13}.

\begin{figure*}[t]
\centering
\includegraphics[width=\linewidth]{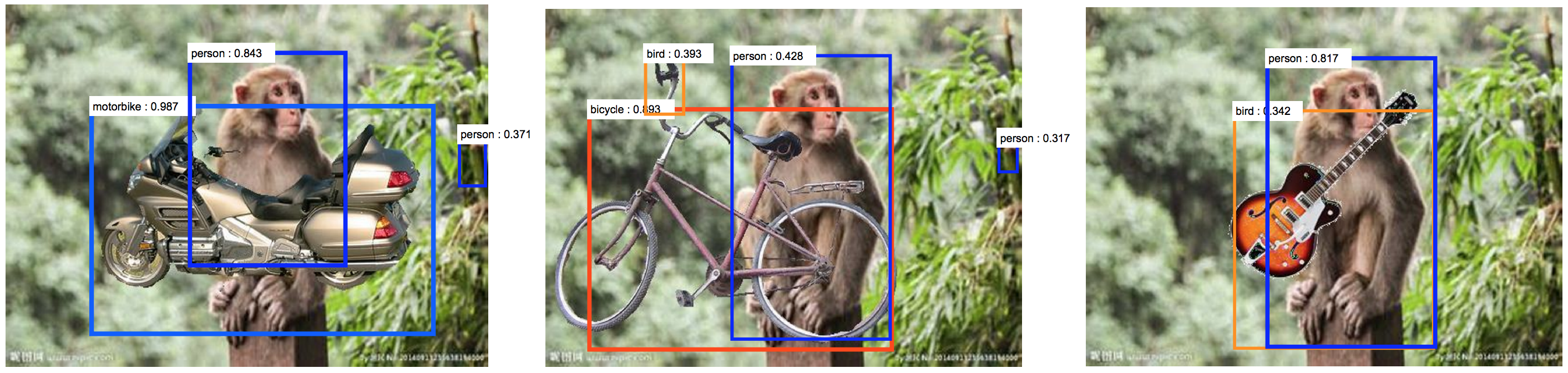}
\caption{Figure taken from \citet{wang2018visual}. Adding occluders cause deep network to fail. Left: The occluding motorbike turns a monkey into a human. Center: The occluding bicycle turns a monkey into a human and the jungle turns the bicycle handle into a bird. Right: The occluding guitar turns the monkey into a human and the jungle turns the guitar into a bird.}
\label{fig:monkey}
\end{figure*}

\subsection{Finding Better Architectures}
\label{sec:nas}

We are now nearly a decade into the third wave of neural network approaches. Looking back, the most important ideas that bring nontrivial improvements to neural network performances almost exclusively belong to the ``architecture'' category, and examples include AlexNet~\citep{DBLP:conf/nips/KrizhevskySH12}, Batch Normalization~\citep{DBLP:conf/icml/IoffeS15}, residual connections~\citep{DBLP:conf/cvpr/HeZRS16}, self-attention~\citep{DBLP:conf/nips/VaswaniSPUJGKP17}, capsules~\citep{DBLP:conf/nips/SabourFH17} etc. Given this progression history, it is certainly possible that a better neural architecture can by itself overcome many of the current limitations.

But what exactly is ``architecture'', anyway? One notable advantage of Deep Nets is that they free researchers from feature engineering, a practice where researchers are responsible for extracting features from the input, and the model is (only) responsible for learning the mapping from the feature to the output. Deep Nets learn layers of feature extractions automatically, so humans can no longer inject their understanding of the problem into the feature extraction process. Instead, the ``architecture'' becomes how humans inject the inductive bias, e.g., by using convolutional nets for images, and recurrent nets for sequence data.

But can Deep Nets learn the architecture as well, by constantly self-improving and making adjustments? This corresponds to an active field of research named Neural Architecture Search. This problem is most naturally solved by reinforcement learning~\citep{DBLP:conf/iclr/ZophL17} or evolutionary algorithms~\citep{DBLP:conf/iccv/XieY17}, but approximations are often needed to reduce the overall cost~\citep{DBLP:conf/eccv/LiuZNSHLFYHM18, DBLP:conf/icml/PhamGZLD18}. In relation to Section~\ref{sec:transfer}, there is also recent evidence~\citep{DBLP:journals/corr/abs-2003-12056} that this process is insensitive to supervision. But overall, we are yet to see truly revolutionary architectures coming out of this process, and as a result, the issues discussed in this paper are still very much relevant. 

\subsection{Addressing Over-Sensitivity to Context}
\label{sec:sensitivity}

\begin{figure*}[t]
\centering
\includegraphics[width=\linewidth]{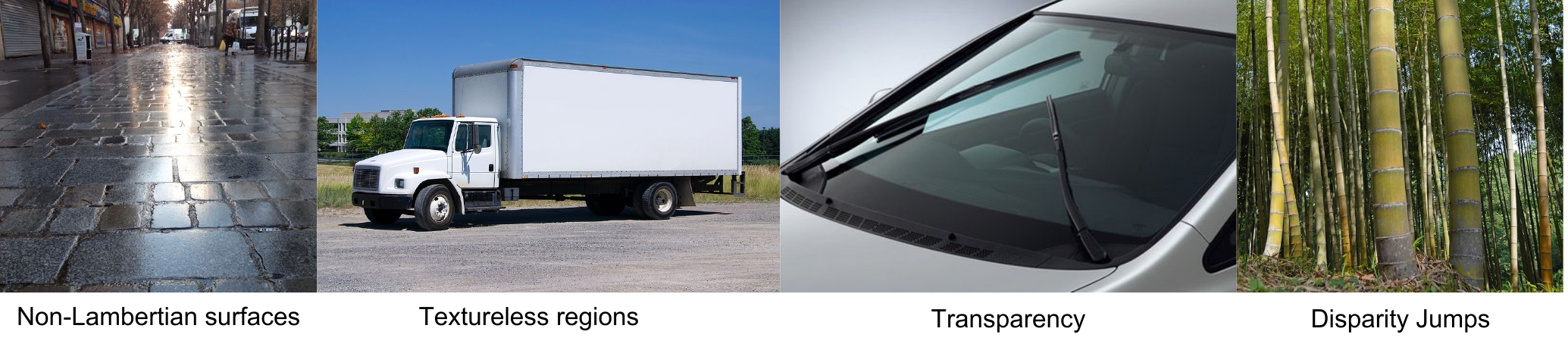}
\caption{Hazardous factors for stereo vision, as identified in \citet{DBLP:conf/iccv/ZendelMHH15}. These challenging scenarios may not occur in real world datasets, and if they do they are not annotated, so relying on synthetic data is a promising alternative. }
\label{fig:hazardous}
\end{figure*}

A more serious challenge to Deep Nets is their over-sensitivity to context. Figure~\ref{fig:monkey} shows the effect of photoshopping a guitar into a picture of a monkey in the jungle. This causes the Deep Net to misidentify the monkey as a human and also misinterpret the guitar as a bird, presumably because monkeys are less likely than humans to carry a guitar and birds are more likely than guitars to be in a jungle near a monkey \citep{wang2018visual}. Recent work gives many examples of the over-sensitivity of Deep Nets to context, such as putting an elephant in a room \citep{DBLP:journals/corr/abs-1808-03305}. Experimental studies show that Deep Nets are less effective than humans when performing object classification on heavily occluded objects~\citep{DBLP:conf/cogsci/ZhuTYPP19, DBLP:conf/wacv/Kortylewski0WZY20}.\footnote{This issue we are describing is in general related to how Deep Nets can take unintended, ``shortcut'' solutions, for example the chromatic aberration noticed in~\citet{DBLP:conf/iccv/DoerschGE15}, or the low level statistics and edge continuity noticed in~\citet{DBLP:conf/eccv/NorooziF16}. In this paper we highlight ``over-sensitivity to context'' as the representative example, for both familiarity and keeping the discussion contained.}

This over-sensitivity to context can also be traced back to the limited size of datasets. For any object only a limited number of contexts will occur in the dataset and so the Deep Net will be biased towards them. For example, in early image captioning datasets it was observed that giraffes only occurred with trees and so the generated captions failed to mention giraffes in images without trees even if they were the most dominant object~\citep{DBLP:journals/aim/ZitnickAAMBP16}. 

Observe that the limited size of datasets is a common theme when we consider the current limitations of Deep Nets. Recall that we already mentioned how synthetic data could be used, see Figure~\ref{fig:unrealcv}, to show that Deep Nets trained on ImageNet could not recognize objects from some viewpoints. An advantage of synthetic data is that it enables us to generate, in principle, an infinite amount of images and hence to systematically explore the effect of varying factors like viewpoint and material properties, e.g., see  \citet{DBLP:conf/eccv/QiuY16, DBLP:conf/cvpr/AlcornLGWMKN19}. Similarly synthetic data can be used to systematically vary hazardous factors for stereo vision (those factors like specularity which are known to cause stereo algorithms to fail; see Figure~\ref{fig:hazardous}) enabling researchers to characterize the sensitivity of stereo algorithms to these factors \citep{DBLP:conf/3dim/ZhangQCHY18}. Hence synthetic datasets offer the possibility of generating as much data as is required to systematically study the sensitivity of Deep Nets to the nuisance factors, like viewpoint and radiosity, which arrive in reality (provided the synthetic datasets are realistic enough to accurately represent real world images).  

The difficulty of capturing the enormous varieties of context, as well as the need to explore the large range of nuisance factors, is highly problematic for data driven methods like Deep Nets. It seems that ensuring that the networks can deal with all these issues will require datasets that are arbitrarily big, which raises enormous challenges for both training and testing datasets. We will discuss these issues next.

%The ``adversarial examples'' discussed above are defined in a rather constrained setting, i.e. the magnitude of the modification, measured in $L_2$ or $L_\infty$, is very small. However, we argue that this distance measure is rather artificial, and the more general criterion should be: the prediction ought to remain the same as long as the content is preserved. To see this, imagine shifting an image to its left by one pixel. This modification will hardly affect human perception, but the induced distance will almost certainly exceed the commonly adopted threshold. Similar views are discussed in more depth in \citet{DBLP:journals/corr/abs-1807-06732}. 

\section{The Combinatorial Explosion: When Big Datasets Are Not Enough}
\label{sec:explosion}

Deep Nets are trained and evaluated on large datasets which are intended to be representative of the real world. But, as discussed earlier, datasets are biased and Deep Nets can fail to generalize to images outside the datasets they were trained on, can make mistakes on rare events that occur rarely within the datasets (but which may have disastrous consequences, such as running over a baby or failure to detect a cancerous tumor), and are also sensitive to adversarial attacks and changes in context. None of these problems are necessarily deal-breakers for the success of Deep Nets and they can certainly be overcome for certain visual domains and tasks. But we argue that these are early warning signs of a problem that will  arise as vision researchers attempt to use Deep Nets to address increasingly complex visual tasks in unconstrained domains. Namely, that in order to deal with the combinatorial complexity of real world images the datasets would have to become exponentially large, which is clearly impractical. 
This forces us to address two new problems: (I) How can we efficiently test these algorithms to ensure that they work in these enormous datasets if we can only test them on a finite subset? (II) How can we train algorithms on finite sized datasets so that they can perform well on the truly enormous datasets required to capture the combinatorial complexity of the real world?

\subsection{The Combinatorial Explosion}

\begin{figure*}[t]
\centering
\includegraphics[width=0.85\linewidth]{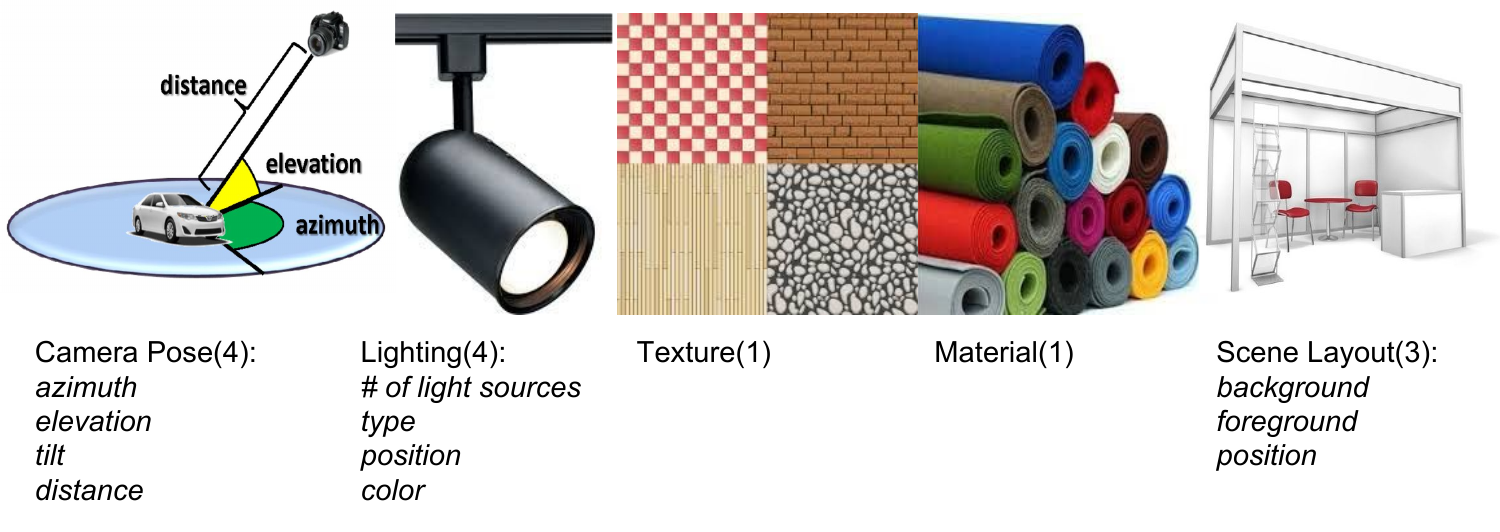}
\caption{An illustration of combinatorial explosion. We consider the (already simplified) rendering process of one object. It involves choosing the camera pose, lighting condition, object texture, etc: a total of (merely) $13$ parameters. If we allow 1,000 different values for each parameter, then we obtain a total of $10^{39}$ different images. This is way beyond the size of any dataset, as well as the number of images humans see per year.}
\label{fig:qi}
\end{figure*}

We now take a step backward and reconsider the problem of vision and the underlying assumptions of our current paradigm for evaluating performance of vision algorithms. As mentioned earlier, in the late 1990s and early 2000s the computer vision community converged on a paradigm where: fixed datasets are collected to approximate the data (and label) distribution; algorithms are trained and tested on this same distribution. This had many advantages since, for the first time it was possible to evaluate and compare the performance of different algorithms. This paradigm  provided a seemingly objective method for evaluating algorithms and to benchmark the  progress of the vision community. It also had rigorous mathematical foundations, such as Probably Approximately  Correct (PAC) theory (also known as VC theory) but since computer vision is largely a pragmatic engineering field it is doubtful how much vision researchers are influenced by this theory (or have studied the underlying theorems and the assumptions that they are based on). We argue that it is time to \emph{rethink this paradigm}. The basic assumptions are that the datasets contain representative, and randomly distributed, samples from the underlying problem domain. For computer vision, due to the infinite space of images, the different image domains, and most fundamentally combinatorial complexity of visual scenes, these assumptions may be too ideal.

So \emph{what do we mean by combinatorial complexity?}
We argue that the most fundamental challenge of vision is that visual scenes, and hence images, can be constructed in a combinatorial number of different ways. It helps to consider this from the perspective of using computer graphics to render the image of a single object. It is straightforward (see Figure~\ref{fig:qi}) to specify a computer program with $13$ parameters that can render images of a single object from different viewpoints, under different illuminations, and in a limited number of background scenes. If we allow 1,000 different values for each parameter we obtain a total of  $10^{39}$ different images, $10^{30}$ times larger than any existing dataset. But this is for a single object. Imagine constructing a visual scene by selecting objects from an object dictionary and placing them in different configurations. This can clearly be done in an exponential number of ways. If we take occlusion into account, we can even obtain combinatorial complexity for images of a single object since it can be partially occluded in an exponential number of ways. The situation becomes even worse if we consider video sequences where objects can move and appear or disappear. 

An image corresponds to one combination out of a combinatorial number of possibilities. Since computer vision can be understood as inverse graphics, the output space of computer vision problems is, therefore, also combinatorially large. In practice though, different visual tasks can be viewed as reducing and simplifying this output space. For example, if we restrict ourselves to object classification, the set of classes is estimated at 20,000 which seems manageable. But for many real world tasks we would have to detect object parts, parse objects in their parts, estimate the 3D position of objects, and do this along the temporal dimension. This soon goes back to the combinatorially large output space. The images in our benchmark datasets (which should now seem very small) are like ``random seeds'' that we throw into this space: it is clear that they will only cover an extremely small fraction of ``bins''. 

This raises the issue of whether the datasets can ever be large enough to be representative of the real world domain on which the algorithms are ultimately intended to run. We will ignore the obvious concerns about dataset bias for the moment and consider the case of a finite-sized unbiased dataset. As the ``seeds'' are too sparse and the number of ``covered bins'' is too small, it is easy for the model to latch on to \emph{accidental, spurious associations} \citep{DBLP:conf/acl/GuuPLL17, DBLP:conf/iclr/KaushikHL20} and settle on a decision boundary that is different from ours (and sometimes considered a ``shortcut''). This could help explain the adversarial example phenomenon, in that the number of training images is too small with respect to the entire image space, so that the majority of the volume (including the $\epsilon$ ball around every training image) is unlabeled. As a result, the decision boundary usually decides to ``cut through'' this $\epsilon$ ball, resulting in adversarial examples. The min-max formulation in adversarial training \citep{DBLP:journals/corr/MadryMSTV17} essentially attempts to enlarge the labeled examples from covering $N$ points to covering $N$ $\epsilon$ balls, now with volume $N \epsilon^d$ where $d$ is the dimension of the image. 
But limitations remain. Firstly, there is no guarantee that the algorithm will find all the adversaries within the $\epsilon$ ball. Secondly, this volume is still an infinitesimal fraction of the true space of images and hence results on the dataset may not apply to the real world. 

\emph{Dataset bias} is, of course, a concern that has existed ever since the first datasets were created. Images in the datasets were typically drawn from photographs which alone causes biases since photographers usually pose their shots and align their cameras carefully.\footnote{The first author remembers that when studying text detection for the visually impaired we were so concerned about dataset biases that we recruited blind subjects who would walk the streets of San Francisco taking images automatically (but found the main difference from regular images was that there was a greater variety of angles).} More generally, however, the datasets often have biased content, because the world we live in has biased content! To revisit an example earlier, early image captioning datasets only contained giraffes next to trees~\citep{DBLP:journals/aim/ZitnickAAMBP16}, because after all, giraffes are much more likely to be near trees than airplanes.
The tendency for datasets to be biased can encourage the algorithms to fit the dataset biases, for example by overfitting to the background context, as illustrated in \citet{DBLP:journals/corr/abs-1912-06314}. Although soccer is often played on a soccer field, there are many situations where people are on a soccer field but are not playing soccer. Conversely, as the first author can testify, soccer can be played in the street, in a backgarden, on a beach, or even on a muddy island in the middle of the Amazon rainforest.  

Another concern regarding dataset bias is that typically datasets do not all try to capture the entire input space. Instead, they often correspond to different domains (subspaces) which may have very different properties and hence algorithms trained on on dataset may perform badly on another. For example, the first two datasets for edge detection, Sowerby\footnote{Available from Sowerby Research Centre, British Aerospace} and South Florida \citep{DBLP:conf/cvpr/BowyerKD99}, corresponded to images of the English countryside and indoor images in Florida. Recall that the task of edge detection is to detect targets, e.g., the boundaries of objects and other salient edge structures, in the presence of background. Edge detection was fairly simple for South Florida, because the background contained almost no texture and so classical edge detectors like Canny \citep{DBLP:journals/pami/Canny86a} were very effective. On the other hand, the background in Sowerby contained highly complex texture and so edge detection was considerably harder. It was found \citep{DBLP:conf/cvpr/KonishiYCZ99, DBLP:journals/pami/KonishiYCZ03} that learning-based methods were highly effective on both datasets separately, but algorithms trained on one dataset gave poor performance on the other (interestingly \citet{DBLP:journals/pami/KonishiYCZ03} was able to solve domain transfer for these two datasets because it used a bayesian formulation instead of simply learning a classifier). More recently, adversarial loss \citep{DBLP:conf/cvpr/TzengHSD17,HoffmanTPZISED18} has been used to learn  domain invariant features, where a domain classifier is trained to distinguish the source and target distributions. \citet{DBLP:conf/cvpr/TzengHSD17} proposes a general framework to bring features from different domains closer. \citet{HoffmanTPZISED18,DBLP:conf/cvpr/MurezKKRK18} extend this idea with cycle consistency to improve results. 
More generally, we can extend the idea of domain transfer to include real images, computer graphics images, line drawings, and clip art images. There are algorithms which are able to transfer between these domains~\citep{DBLP:journals/corr/abs-1912-08265} and, it should be noted, that humans are extremely good at domain transfer.

\subsection{Testing Models When Data Is Combinatorial}

How can we evaluate the performance of visual algorithms in light of the issues raised in the previous section? The standard paradigm of training and testing on a finite number of randomly drawn samples has made a huge contribution to the subject and will still be useful when used sensibly, but we argue that it is no longer sufficient and  needs to be replaced by other performance measures that take into account the infinite space of images and their combinatorial complexity. In this section, we sketch a range of complementary strategies for achieving this.

One strategy is to keep the same evaluation paradigm but create a much larger set of benchmark challenges. These challenges could be inspired by the abilities of the human visual system which, as discussed earlier is superior to current computer vision systems in many respects.
Firstly, it performs a multitude of visual tasks at the same time  (e.g., detect objects, parse them into parts, find their boundaries, and estimate their 3D configurations). Secondly, humans can learn very efficiently from small numbers of examples presumably by exploiting prior knowledge and physical properties of the world. Thirdly, the human visual system is robust to viewpoint changes, to novel contexts, partial occlusion (including overlapping objects as in CAPTCHAs), and to pixel-level adversarial image attacks. Fourthly, humans can learn on one image domain, e.g., real images, line drawing, or visual arts, and effortlessly transfer to new domains (often with no supervision).
This is already being pursued by, for example, requiring algorithms to transfer between different domains~\citep{DBLP:journals/corr/abs-2003-04490} and to test on out-of-distribution data~\citep{DBLP:journals/corr/abs-2003-08440}, but there is much more to be done. 

Another strategy is to try to identify the types of stimuli which are difficult for the algorithm and include performance measures which paid attention to the hardest cases. This involves the notion of hard negative mining and error analysis. For example, for object detection on PASCAL it was known that algorithms like deformable part models~\citep{DBLP:journals/pami/FelzenszwalbGMR10} performed very well except on small objects~\citep{DBLP:conf/eccv/HoiemCD12}. It is well known that semantic segmentation algorithms are rewarded based on the number of pixels they classify correctly so they are biased towards detecting large regions and typically perform less well on smaller regions. More recently work on face detection can identify the most challenging cases and develop algorithms that perform well on them~\citep{DBLP:conf/wacv/ZhangSQWWY20}. In some cases, the structure of the visual task makes it possible to identify which types of stimuli would most challenge the algorithms. For example, as mentioned earlier in Section~\ref{sec:sensitivity}, researchers have isolated the hazardous factors which cause stereo algorithms to fail which include specularities and texture-less regions. In such cases it is possible to exploit computer graphics to systematically vary these hazardous factors to determine which algorithms are resistant to them \citep{DBLP:conf/3dim/ZhangQCHY18}. In general, if we understand characteristics of the visual task we could design datasets that target the difficult and challenging cases. But this requires that the challenging cases are, in some sense, low-dimensional so that they can be systematically explored. 

But we argue that even more radical changes in evaluating performance will be needed to deal with combinatorial complexity and the infinite space of images. To obtain computer vision systems that work reliably in the real world we would need to evaluate performance on a large range of tasks and over an infinite set of images. We also need to use alternative measures, such as worst case, instead of relying on average case. This makes sense if the goal is to develop visual algorithms for self-driving cars, or diagnosing cancer in medical images, where failures of the algorithms can have major consequences. Firstly, we generalize the notion of attacks to emphasize the worst case. Secondly, we leverage computer graphics to overcome the limitation of a fixed dataset of limited image samples (or create additional images at run time by generalizing adversarial methods \citep{DBLP:journals/corr/abs-2004-05682}). Thirdly, we adapt the test cases based on the test history, which ``personalizes'' the testing experience and improves sample efficiency (as different models have different weaknesses).
This \emph{Adversarial Examiner} approach is illustrated in \citet{DBLP:conf/aaai/ShuLQY20}. In other words, the images are systematically chosen to probe for the weaknesses of the trained model. This requires, of course, overcoming the challenging problem of how to define strategies which efficiently search over the huge space of images. In \citet{DBLP:conf/aaai/ShuLQY20} we used reinforcement learning and Bayesian optimization, to learn a policy for selecting the sequence of test images. 

\begin{figure*}[t]
\centering
\includegraphics[width=\linewidth]{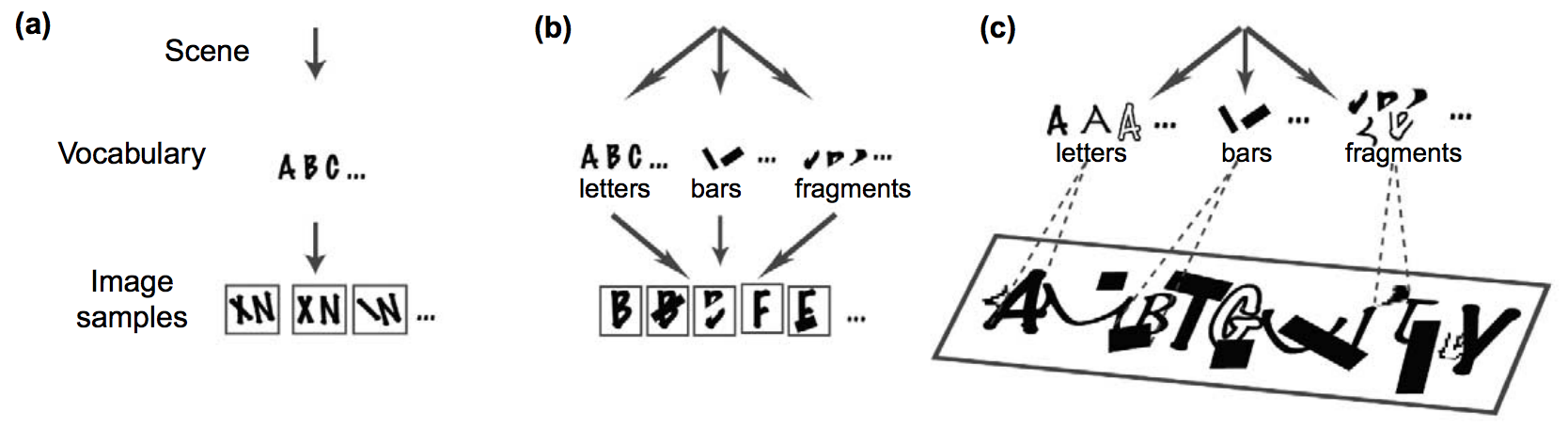}
\caption{Figure taken from \citet{yuille2006vision}. From (a) to (b) to (c), an increasing level of variability and occlusion is used, yet humans can still do inference and correctly interpret the image.}
\label{fig:letters}
\end{figure*}

\subsection{Methods for Overcoming Combinatorial Complexity}

What types of algorithms will be able to deal with combinatorial complexity and perform well on the new types of benchmark challenges that we describe in the previous section? 

It is possible that current Deep Nets, or advanced variants of them, will be sufficient to deal with the challenges. Perhaps some of the apparent limitations of current Deep Nets are because they are smart at taking shortcuts to exploit the limitations of the current datasets and, if faced with harder challenges, they would rise to the occasion? After all, Deep Nets are a rapidly developing and innovating research field. And, given their past successes,  this would be a sensible strategy to pursue.
But we suspect that simply extending current methods will not be enough. As our ultimate goal is to match or surpass our own biological vision, it is natural to get inspiration from studies of it.  Humans see at most roughly $10^9$ images every year (assuming 30 images per second) which is big, but not combinatorial, and humans certainly learn without detailed supervision. Instead they acquire visual skills over a period of years in a stereotyped sequence where different skills develop at different times \citep{arterberry2016development} and are not passive observers but instead interact with the world behaving like ``baby scientists'' who seek to understand the causal structure of the world \citep{gopnik2004theory}. Human learning appears to be analogous to, but infinitely better than,  classic AI systems which systematically build up knowledge representations about the world \citep{DBLP:books/daglib/0023820} and solve new problems by exploiting this knowledge, e.g., a human could learn the structure of a new car from a brief glance because of previously acquired knowledge about cars. In addition, humans can perform a huge range of visual tasks and have other properties: see Section~\ref{sec:cogsci}.

What will computer vision systems need in order to achieve these abilities? It has long been speculated that they will need \emph{generative} models which have the capability of synthesizing real world images from representations of the world and perform analysis by synthesis  \citep{grenander1993general, mumford1994pattern, DBLP:conf/iccv/TuCYZ03, DBLP:journals/ftcgv/ZhuM06, mumford2010pattern}. This can be seen as mathematical instantiation  of the ideas expressed by Gregory \citep{gregory1973eye} and can be given a Bayesian formulation in terms of priors and likelihood functions. These priors, which can be extended to include knowledge about intuitive physics~\citep{battaglia2013simulation} enable humans to imagine and predict. Generative models moreover offer the potential to address many of the challenges discussed earlier. The success and maturity of computer graphics shows that we now have the tools for synthesizing increasingly realistic images from a representation of the three-dimensional world. Computer vision, however, is faced with the much harder task of inverse inference where, given an image, we have to decide which is the most likely state of the world. In view of the combinatorial complexity of visual scenes this is an extremely challenging problem.

One attractive strategy is develop generative models which are \emph{compositional}. Compositionality is a general principle which can be described poetically as ``an embodiment of faith that the world is knowable, that one can tease things apart, comprehend them, and mentally recompose them at will''. The key assumption is that structures are composed hierarchically from more elementary substructures following a set of grammatical rules. This suggests that the substructures and the grammars can be learned from finite amounts of data but will generalize to combinatorial situations. Compositional models require structured representations which make explicit their structures and substructures which enables them to do multiple tasks (e.g., detecting objects, object parts, and object boundaries) with the same underlying representation \citep{DBLP:conf/nips/ChenZLYZ07}. Compositional models offer the ability to extrapolate beyond data they have seen, to reason about the system, intervene, do diagnostics, and to  answer many different questions based on the same underlying knowledge structure \citep{pearl2009causality}. To quote Stuart Geman ``the world is compositional or God exists'', since otherwise it would seem necessary for God to handwire human intelligence \citep{geman2007compositionality}.

%Recent experiments \citep{ullman2016atoms} suggest that humans can interpret images unambiguously provided they are above a critical size (which depends on the image content) and additional context is unnecessary. 

We can illustrate these ideas by some toy-world examples, shown in Figure~\ref{fig:letters}, where images are created in terms of basic vocabularies of elementary components (these examples were developed in \citet{yuille2006vision}).  The three panels show microworlds of increasing complexity from left to right. For each microworld there is a grammar which specifies the possible images as constructed by compositions of the elementary components. In the left panel the elementary components are letters which do not overlap, and so interpreting the image is easy. The center and right panels are generated by more complicated grammars -- letters of different fonts, bars, and fragments which can heavily occlude each other. Interpreting these images is much harder and seems to require the notion that letters are composed of elementary parts, that they can occur in a variety of fonts, and the notion of ``explaining away'' (to explain that parts of a letter are missing because they have been occluded by another letter). 

\begin{figure*}[t]
\centering
\includegraphics[width=0.9\linewidth]{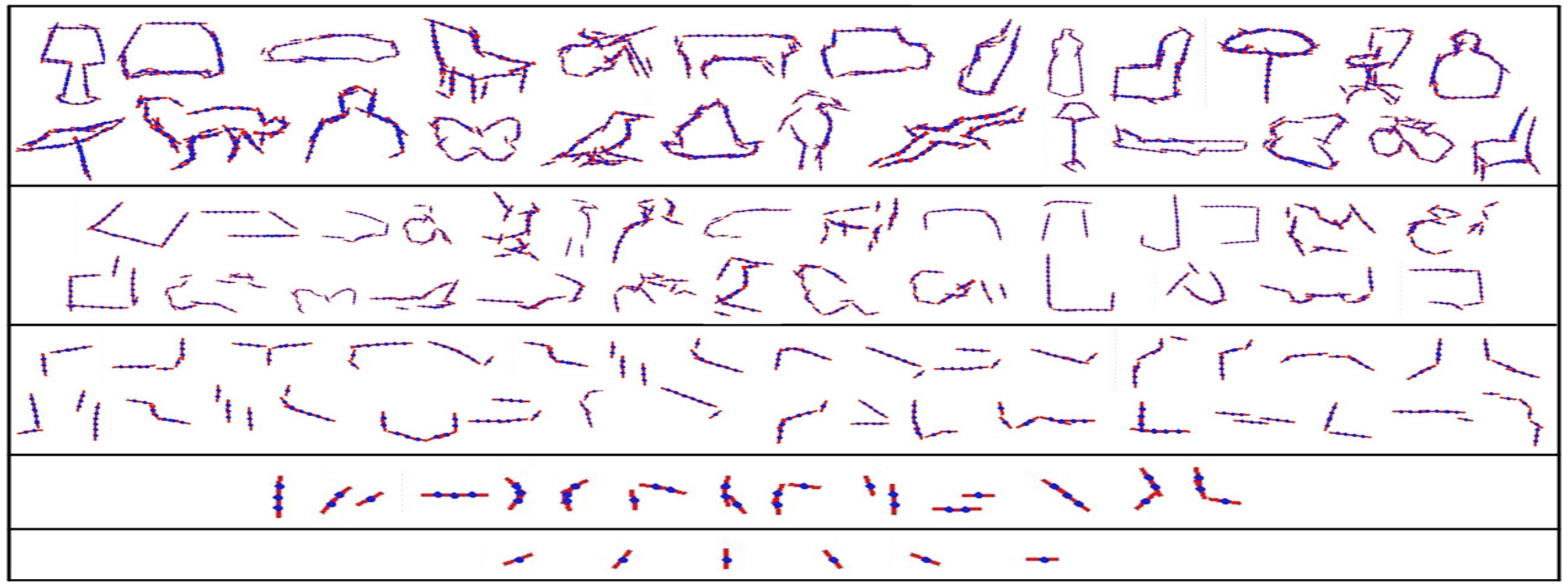}
\caption{
% Top: 
Figure taken from \citet{DBLP:conf/cvpr/ZhuCTFY10}. Mean shapes from Recursive Compositional Models at different levels. This hieararchy was learned in an unsupervised manner.
% Bottom: Figure taken from \citet{DBLP:conf/cvpr/WangY15}. One learned mixture with corresponding landmark localization of horse images.
}
\label{fig: composition}
\end{figure*}

The third microworld in Figure~\ref{fig:letters} is an example of a combinatorially large dataset since images are constructed by selecting objects from a dictionary and placing them at random while allowing for occlusion. This microworld is essentially the same as CAPTCHAs which can be used to distinguish between humans and robots. Interestingly, work on CAPTCHAs  \citep{george2017generative} show that compositional models which represent objects in terms of compositions of elementary tokens and  factorize geometry and appearances can perform well on these types of datasets (note that this method was applied only to text and digits and has not been extended to objects in natural images and has not been tested on standard datasets). Their inference algorithm involves bottom-up and top-down processing \citep{DBLP:conf/iccv/TuCYZ03} which enables the algorithm to ``explain away'' missing parts of the letters and to impose ``global consistency'' of the interpretation to remove ambiguities. Intuitively, part detectors  make bottom-up proposals for letters which can be validated or rejected in the top-down stage. By contrast, Deep Nets performed much worse on these datasets. Presumably because, unlike compositional models, they cannot capture the underlying generative structure of the domain and extrapolate outside their training dataset. Since the microworld is combinatorially large, it will not be possible to train Deep Nets on enough data to guarantee good performance on the entire dataset.
Other non-visual examples illustrate the same points. A recent example is when researchers \citep{DBLP:conf/icml/SantoroHBML18} tried to train standard Deep Nets to do IQ tests. The task requires finding composition of meaningful rules/patterns (distractors may be present) within 8 given images in a $3 \times 3$ grid, and the goal is to fill in the last missing image. Not surprisingly, Deep Nets do not generalize well. 
Theoretical studies, e.g., \citet{DBLP:journals/jmlr/YuilleM16}, suggest that compositional models are well suited for dealing with complexity by sharing parts and using hierarchical abstraction.

The design of Deep Nets certainly has the idea of compositionality and hierarchy built in. But there are several aspects that deviate from the descriptions above. Firstly, the nodes in Deep Nets are \emph{not symbolic}, which makes abstraction and interpretability a lot more challenging. Secondly, the Deep Nets that we most frequently use are \emph{not generative}. Deep Nets have been applied to generating images using Generative Adversarial Networks (GANs)~\citep{DBLP:conf/nips/GoodfellowPMXWOCB14} but except for conditional GANs~\citep{DBLP:journals/corr/MirzaO14}, these lack the types of semantic structure that we believe will be needed. Thirdly, Deep Nets usually have a fixed architecture, which can make compositions a lot \emph{less flexible}. In this sense, Neural Module Networks \citep{DBLP:conf/cvpr/AndreasRDK16} are more promising than static Deep Nets, in that the dynamic architectural layout may be flexible enough to capture some meaningful compositions. In fact, we recently verified that the individual modules indeed perform their intended functionalities (e.g. \texttt{AND}, \texttt{OR}, \texttt{Filter(red)} etc) after joint training \citep{DBLP:conf/cvpr/LiuLBY19}. But we have yet to see them being successfully applied on a wider range of computer vision tasks. 
Finally, we comment that researchers are trying to marry desirable aspects of both sides, e.g., by being generative but on the convolutional features \citep{DBLP:journals/corr/abs-2003-04490}. 

Compositional models have many desirable theoretical properties, such as being interpretable, and the ability to be generative so they can be sampled from. This means that, in principal, they know everything about the object (or whatever entity is being modeled) which makes them easier to diagnose, and hence harder to fool, than black box methods like Deep Nets. But learning compositional models is hard because it requires learning the building blocks and the grammars (and even the nature of the grammars is debatable). It is an open question how the structure of the compositionality ought to emerge, though some studies suggest evolution could prompt modular \citep{clune2013evolutionary} and hierarchical \citep{DBLP:journals/ploscb/MengistuHMC16} structures. There has, however, been some limited success in learning hierarchical dictionaries starting from basic elementary tokens like edges \citep{DBLP:conf/cvpr/ZhuCTFY10}: see Figure~\ref{fig: composition}. Also, putting distributions on images is challenging in general, with a few exceptions like faces, letters, and regular textures \citep{DBLP:conf/iccv/TuCYZ03}.

%For example, it is necessary to have rich appearance models which can factorize between shape and appearance, see \citet{DBLP:conf/cvpr/WangY15}. But the increasing realism of virtual worlds constructed using computer graphics tools, e.g., see Figure~\ref{fig:unrealcv}, suggests we may soon have realistic generative models which make analysis by synthesis possible.

More fundamentally, dealing with the combinatorial explosion requires learning causal models of the 3D world and how these generate images. Studies of human infants suggest that they learn by making causal models that predict the structure of their environment including naive physics. This causal understanding enables learning from limited amounts of data and performing true generalization to novel situations. This is analogous to contrasting Newton's Laws, which gave causal understanding with a minimal amount of free parameters, with the  Ptolemaic model of the solar system gave very accurate predicts but required a large amount of data to determines its details (i.e. the epicycles). Ptolemaic, Copernican, and Kelper all describe the Solar System fairly well. And Kelper may be preferred because it is simplest (though Kepler had more than three laws and Newton had to decide which laws were correct and not redundant). But crucially, Newton's Laws are more causal because they not only explain the solar system but also why apples fall to the ground (and many other gravitational phenomena). Newton could also predict what would happen if a rogue planet, or gigantic spacecraft, entered the solar system but Ptolomy, Copernicus and Kepler could not.

\section{Performance Evaluation and Data Bias Outside Academia}
\label{sec:outside}

This opinion paper has concentrated on academic issues and argued that we need more advanced performance measures when evaluating deep networks and other computer vision algorithms. But in this section we step outside the computer vision academic community and make two additional points:  (I) Performance evaluation, and in particular rethinking evaluation metrics, matters outside academia. (II) Data bias exists, causes serious real world problems, and we need better algorithms that can learn to be fair even when trained on biased data.

In the last ten years computer vision, almost entirely due to the success of deep networks, has become well known to the general public. It is a vital part of the artificial intelligence technology which has a huge range of practical applications that have the potential to greatly change human society. Countries and even continents are making strategic plans for artificial intelligence which involve investing many billions, and maybe even trillions, of dollars. This means that the performance of vision algorithms, and how we evaluate them, is no longer of purely academic concern. Like  more mature technologies --- e.g., trains, bridges, cars, and aircraft --- we need performance measures that are adequate to deal with the complexities of the real world (in particular, situations which can cause serious harm) and methods for diagnosing failure modes. We argue that current performance measures are not adequate and lead, on the one hand, to unrealistic expectations, and on the other to dangers that the algorithms will be used in ways that will adversely affect humans. 

Some of the dangers are fairly obvious since failure of the systems could result in loss of life. For example, would you trust your life to a robotaxi if it is raining? Depth sensors like lidar are problematic in rainy conditions while, as computer vision researchers should know, rain causes specularities which are known to be harzardous factors for depth estimation by binocular stereo \citet{DBLP:conf/iccv/ZendelMHH15} and this has only partially been quantified \citep{DBLP:conf/3dim/ZhangQCHY18}
 due to the difficulty of annotating specularities in datasets.  And would you trust a self-driving car to avoid driving over bicyclists late at night or, even worse, to fail to stop in time if there was a baby sitting in the road ahead of you? These are highly dangerous situations which may not even appear in the training and testing datasets of your algorithm (no matter how large your dataset). Increasing your dataset by adding all these ``corner cases'' is problematic due to the compositional complexity issues we discussed in the previous section. For these types of reasons the automated car industry now realizes that it will take longer to build self-driving cars than everyone thought.\footnote{\url{https://www.caranddriver.com/features/a32266303/self-driving-cars-are-taking-longer-to-build-than-everyone-thought/}} Similarly, although the use of deep networks to perform medical imaging tasks such as early detection of cancerous tumors has enormous potential, there are also significant dataset bias issues to be addressed. Computer Tomography (CT) images at different hospitals are taken using different types of scanners with a variety of imaging protocols (e.g., the scans being taken at specific times after the patient has invested specific quantities of water or dye). Hence algorithms that may work almost perfectly on a benchmark dataset of CT scans obtained at one hospital may fail badly on CT scans at other hospitals, potentially leading to misdiagnosis and risk of a painful death to the patient. Moreover, cues for detecting cancer can be very subtle and there are rare corner cases which only highly skilled radiologists can detect. Other major problems arise when deep networks are used for tasks involving faces. Few datasets, if any, represent the immense variation of facial appearance which arise due to gender and the enormous variety of demographics. This can result, for example, in people in under-represented demographics being unfairly treated by deep networks trained on biased datasets \citep{buolamwini2018gender}. Even more concerning, it has been shown that certain biases in datasets can get amplified by the deep networks even if datasets are balanced \citep{wang2019balanced}. This is disturbingly reminiscent of historical problems where medical research focused too much on diseases of men\footnote{\url{https://www.ncbi.nlm.nih.gov/books/NBK210143/}} (perhaps particularly of rich men), and where IQ tests were badly misused for immigration and sterilization \citep{bashford2010oxford}.

To avoid, or at least mitigate, these problems we argue that the computer vision community should think much more carefully about the performance measures that we currently use for testing our algorithms and work on developing new performance measures better suited to the combinatorial complexity of the real world. We should also all be aware of data bias and the many serious problems that it can cause.  Our personal opinion is that eliminating bias in datasets, even if this is completely practical, will not be enough. Instead we argue that addressing these issues will also require developing more advanced algorithms which can generalize beyond the biases in the data and which are interpretable, explainable, and diagnosable.  Finally, even if we can avoid the issues described here we do not want computer vision to be used to support the type of dystopian societies described in George Orwell's book 1984 or in Terry Gilliam's film Brazil (where a fly jammed in a teleprinter causes an innocent person to be identified as a terrorist). These political and ethical issues, however, are out of scope of this paper but our opinion is that researchers should be thinking seriously about them. In general we are optimistic and agree with others\footnote{\url{https://www.wired.com/story/done-right-ai-make-policing-fairer/}} that the benefits of computer vision, if wisely used, significantly outweigh the disadvantages. But we argue that realistic performance measures are a necessary prerequisite.

\section{Conclusion \label{sec:conclusion}}

This opinion piece has been motivated by discussions about Deep Nets with researchers in several disciplines including researchers in both computer and biological vision. We have tried to strike a balance which acknowledges the immense success of Deep Nets but which does not get carried away by the popular excitement surrounding them. But we stress that  views expressed in the paper are our own and do not necessarily reflect those of the computer vision community. Our references are far from exhaustive but instead give entry points into the vast literature on these topics. 

A few years ago Aude Oliva and the first author co-organized a NSF-sponsored workshop on the Frontiers of Computer Vision (MIT CSAIL, August 21-24 2011). The meeting encouraged frank exchanges of opinion and, in particular, there was enormous disagreement about the potential of Deep Nets for computer vision. But a few years later, as Yann LeCun predicted, everybody is using Deep Nets. Their successes have been extraordinary and have helped vision become much more widely known, dramatically increased the interaction between academia and industry, lead to application of vision techniques to a large range of disciplines, and have many other important consequences. But despite their successes there remain enormous challenges which must be overcome before we reach the goal of general purpose artificial intelligence and understanding of biological vision systems. 

Our major concern is that the space of natural images is combinatorially large which raises concerns about how we evaluate deep nets and other vision algorithms. Current performance measures involve finite-sized testing sets which, in light of the combinatorial complexity may be poor predictors of how the algorithms will work in the complexity of real world conditions. Hence claims that ``object recognition is a solved problem and we know this because performance on ImageNet is so good that the competition has been retired''\footnote{Quote during a public talk by a West Coast Professor who, perhaps coincidentally, had a start-up company.} should be treated with caution and we should keep in mind that ``an algorithm is only as good as the dataset it has been tested on and the performance measures used''. The combinatorial complexity of vision requires rethinking how the performance of computer vision algorithms ought to be evaluated.

We suggest that they should be tested by tougher challenges, inspired by the abilities of the human visual system, and perhaps by using adversarial examiners to probe for their weaknesses by allowing images to be modified at test time. We question whether Deep Nets will be sufficient to address these challenges and argue that methods that are compositional, generative, and combine signal and symbolic processing will be needed. From the learning side, we suggest that novel approaches to learning should be developed based on findings from the developmental literature. We should, once again, acknowledge the great progress achieved by Deep Nets. It is a testimony to their achievements that we can even think seriously about how to address the challenge of the combinatorial complexity of visual images. But finally, we should point out that since Deep Nets are increasingly being applied to real world problems improved performance evaluation, and dealing with dataset bias, will not only benefit academic research but is also of vital importance as computer vision takes its  central place in the artificial intelligence technological revolution.

%Several of our concerns parallel those mentioned in recent critiques of Deep Nets \citep{DBLP:journals/cacm/Darwiche18, DBLP:journals/corr/abs-1801-00631}. 

%In particular dealing with the combinatorial explosion as researchers address increasingly complex visual tasks in real world conditions. While Deep Nets, and other big data methods, will surely be part of the solution we believe that we will also need complementary approaches which can build on their successes and insights. 

%KEY POINTS: model-based versus curve fitting (Darwiche) -- neglect of non-DP approach (particularly in teaching) -- history --over-promise of expert systems -- the need for representations and true compositionality -- the metrics of performance are misleading (datasets) --increasingly sophisticated statistical and optimization techniques -- automation and not AI --``bullied by success''.

%Overfit to background context in training sets and lack of transfer. Using human-abilities to challenge AI (what can humans do? Often better answered by being a human than by studying the scientific literature). The need for better performance metrics. Ultimately we need to abstract and get models.

\begin{acknowledgements}
This work was supported by the Center for Brains, Minds and Machines (CBMM), funded by NSF STC award CCF-1231216 and ONR N00014-15-1-2356. We thank Kyle Rawlins, Tal Linzen, Wei Shen, and Adam Kortylewski for providing feedback and Weichao Qiu, Daniel Kersten, Ed Connor, Chaz Firestone, Vicente Ordonez, and Greg Hager for discussions on some of these topics. We thank the reviewers for some very helpful feedback which greatly improved the paper.
\end{acknowledgements}

% BibTeX users please use one of
\bibliographystyle{spbasic}      % basic style, author-year citations
\bibliography{refs}   % name your BibTeX data base

\end{document}